\documentclass[sigconf,screen]{acmart}
\usepackage{subfigure}
\usepackage{bm}

\usepackage{amssymb}
\newcommand\blfootnote[1]{%
  \begingroup
  \renewcommand\thefootnote{}\footnote{#1}%
  \addtocounter{footnote}{-1}%
  \endgroup
}
\AtBeginDocument{%
  \providecommand\BibTeX{{%
    \normalfont B\kern-0.5em{\scshape i\kern-0.25em b}\kern-0.8em\TeX}}}

\copyrightyear{2021}
\acmYear{2021}
\setcopyright{acmcopyright}
\acmConference[MM '21]{Proceedings of the 29th ACM International Conference on Multimedia}{October 20--24, 2021}{Virtual Event, China}
\acmBooktitle{Proceedings of the 29th ACM International Conference on Multimedia (MM '21), October 20--24, 2021, Virtual Event, China}
\acmPrice{15.00}
\acmDOI{10.1145/3474085.3475467}
\acmISBN{978-1-4503-8651-7/21/10}

\acmSubmissionID{mfp1676}

\begin{document}
\fancyhead{}

\title{DPT: Deformable Patch-based Transformer for Visual Recognition}

\author{Zhiyang Chen$^{1,2}$, Yousong Zhu$^{1}$, Chaoyang Zhao$^{1\ast}$, Guosheng Hu$^{3}$, Wei Zeng$^{4}$, Jinqiao Wang$^{1,2}$, Ming Tang$^{1}$}
\affiliation{
    \institution{$^1$National Laboratory of Pattern Recognition, Institute of Automation, Chinese Academy of Sciences, Beijing, China}
    \institution{$^2$School of Artificial Intelligence, University of Chinese Academy of Sciences, Beijing, China}
    \institution{$^3$AnyVision, Belfast, UK}
    \institution{$^4$Peking University, Beijing, China}
    \city{}
    \country{}
}
\email{{zhiyang.chen, yousong.zhu, chaoyang.chao, jqwang, tangm}@nlpr.ia.ac.cn}
\email{huguosheng100@gmail.com, weizeng@pku.edu.cn}

\renewcommand{\shortauthors}{Zhiyang Chen, et al.}

\begin{abstract}
  Transformer has achieved great success in computer vision, while how to split patches in an image remains a problem. 
  Existing methods usually use a fixed-size patch embedding which might destroy the semantics of objects. To address this problem, 
  we propose a new Deformable Patch (DePatch) module which learns to adaptively split the images into patches with different positions and scales in a data-driven way rather than 
  using predefined fixed patches. In this way, our method can well preserve the semantics in patches.
  The DePatch module can work as a plug-and-play module, which can easily be incorporated into different transformers to achieve an end-to-end training.  We term this DePatch-embedded transformer as Deformable Patch-based  Transformer (DPT) and conduct extensive evaluations of DPT on image classification and object detection. Results show DPT can achieve 81.9\% top-1 accuracy on ImageNet classification, and 43.7\% box mAP with RetinaNet, 44.3\% with Mask R-CNN on MSCOCO object detection. Code has been made available at: \url{https://github.com/CASIA-IVA-Lab/DPT}.
\end{abstract}



\maketitle
\blfootnote{$\ast$ Corresponding author.}

\begin{figure}[h]
    \subfigure[Original patch embedding module.]{
      \includegraphics[width=1\linewidth]{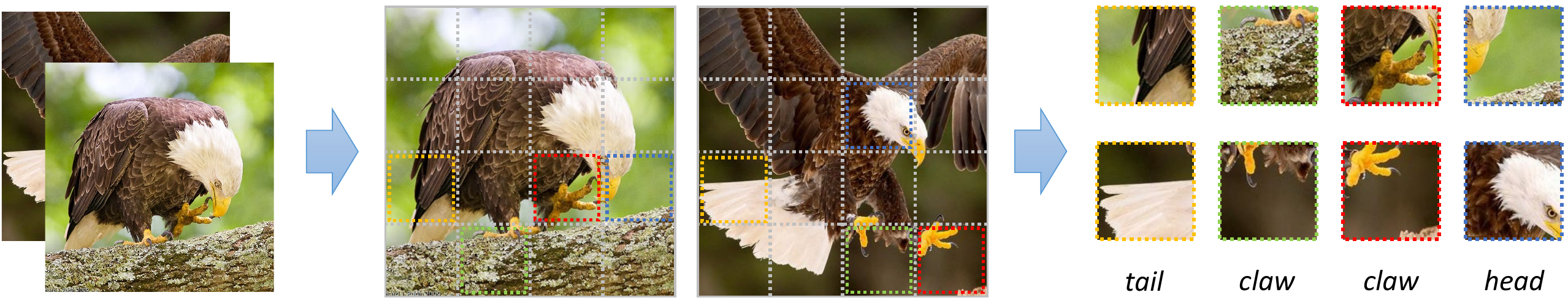}
      \label{fig:patch_old}
    }
    \subfigure[DePatch.]{
      \includegraphics[width=1\linewidth]{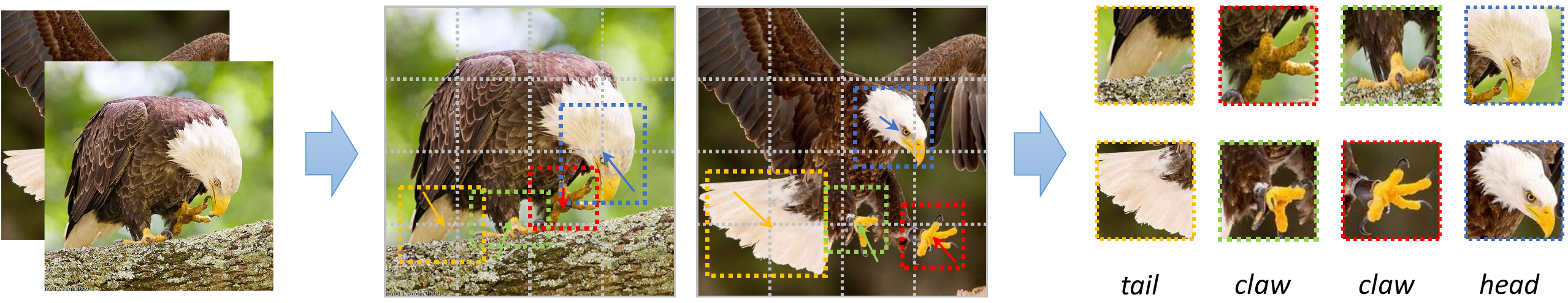}
      \label{fig:patch_new}
    }
  \caption{An example of vanilla patch splitting and our deformable way. (a) Original patch embedding module divides the image in a fixed way. It sometimes destroys the semantics of objects. (b) Our DePatch splits the image into patches in a deformable way with learnable positions and scales. (Better viewed in color)}
  \label{fig:introduction}
  \Description{Introduction figure.}
\end{figure}

\section{Introduction}

Recently, transformer \cite{transformer} has made significant progress in natual language process (NLP) and speech recognition. It has gradually become the prevailing method for sequence modeling tasks. Inspired by this, some studies have successfully applied transformer in computer vision, and achieved promising performance on image classification \cite{deit,swin}, object detection \cite{detr, deformabledetr}, and semantic segmentation \cite{setr}. Similar to NLP, transformer usually divides the input image into a sequence of fixed-size patches (e.g. $16\times 16$) \cite{vit,deit,pvt,t2tvit}, and models the context relationships between different patches through multi-head self-attention. Compared to convolution neural networks (CNNs), transformer can effectively capture long-range dependency inside the sequence, and the features extracted contain more semantic information.

Although transformer has tasted sweetness in vision tasks, there are still many aspects to be improved. DeiT \cite{deit} exploits data augmentation and knowledge distillation to learn visual transformer in a data-efficient manner. T2T-ViT \cite{t2tvit} decomposes the patch embedding module by recursively aggregating neighboring tokens for better local representation. TNT \cite{tnt} maintains a fine-grained branch to better model local details inside a patch. PVT \cite{pvt} transforms the architecture into four stages, and generates feature pyramid for dense prediction. 
These works use a \emph{fixed-size} patch embedding under an implicit assumption that the fixed image split design is suitable for all images. 
However, such `hard' patch split may bring two problems: (1) Collapse of local structures in an image. As shown in Figure~\ref{fig:patch_old}, a regular patch ($16\times 16$) is always hard to capture the complete object-related local structure, since objects are with various scales in different images. (2) Semantic inconsistency across images. The same object in different images might have different geometric variations (scale, rotation, etc). The fixed way of splitting images will potentially capture inconsistent information for one object in different images. 
As discussed, these fixed patches can potentially destroy the semantic information, leading to degraded performance.

To address the aforementioned problems, in this paper, we propose a new module, called DePatch, which divides images in a deformable way. In this way, we can well preserve the semantics in one patch, reducing the semantics destruction caused by image splitting. To achieve this, we learn the offset and scale of each patch in feature map space. The offset and scale are learned based on the input feature map and are generated for each patch as illustrated in Figure~\ref{fig:patch_new}.
The proposed module is lightweight and introduces a very small number of parameters and computations.
More importantly, it can work as a plug-and-play module which can easily be incorporated into other transformer architectures. A transformer with DePatch module is named  Deformable Patch-based Transformer, DPT. In this work, we integrate DePatch module to Pyramid Vision Transformer (PVT) \cite{pvt} to verify its efficacy since PVT achieves state-of-the-art performance in pixel-level prediction tasks like object detection and semantic segmentation.  
With deformable patch adjustment, DPT generates complete, robust and discriminative features for each patch based on local contextual structures. Therefore it can not only achieve high performance on classification tasks, but also outperform other methods on tasks which highly depend on local features, e.g. object detection. 
Our method achieves 2.3\% improvements on ImageNet classification, and improves box mAP by 2.8\%/3.5\% with RetinaNet and Mask R-CNN framework for MSCOCO object detection compared to its conterpart, PVT-Tiny.

Our main contributions can be summarized as:

\begin{itemize}
\item We introduce a new adaptive patch embedding module,  DePatch. 
DePatch can adjust the position and scale of each patch based on the input image, and effectively preserve semantics in one patch, reducing semantics destruction caused by image splitting.  
\item Our DePatch is lightweight and can work as a plug-and-play module integrated into different transformers, leading to a Deformable Patch-based Transformer (DPT). In this work, we incorporate DePatch into PVT to verify the efficacy of DPT.

\item We conduct extensive experiments on image classification and object detection. For example, our module improves top-1 accuracy by 2.3\% on ImageNet classification, and also gains 2.8\%/3.5\% improvements for both RetinaNet and Mask R-CNN detectors under the tiny configuration. 
\end{itemize}

\section{Related Work}
\subsection{Vision Transformer}
Transformer \cite{transformer} has been the mainstream approach for NLP tasks. It uses self-attention to capture long-range dependence within the whole sequence, and achieves state-of-the-art performance. This idea is applied into computer vision firstly by Non-Local block and its variants \cite{nonlocal,gcnet,ccnet}. Recently, there appear a large amount of works building pure vision transformers without convolution layers. ViT \cite{vit} is as far as we know the first work in this trend. It achieves comparable results with traditional CNN architectures with the help of large training data. DeiT \cite{deit} uses complex training schedules and knowledge distillation to improve performance trained on ImageNet only.

Current works focus on combining the advantages of transformer and CNN in order to capture better local information. This object is obtained by combining convolution blocks and self-attention layers together \cite{botnet,cpvt}, maintaining high-resolution feature maps \cite{tnt,pvt,swin}, adding parameters biased for locality \cite{smca,convit} or re-designing the brute-force patch-embedding module \cite{t2tvit}. Though large improvements achieved, most architectures split the input image with a fixed pattern, without awareness of the input content and geometric variations. Our DPT can modify the position and scale of each patch in an adaptive way. To the best of our knowledge, our model is the first vision transformer that do patch splitting in a data-specific way.

\subsection{Deformable-Related Work}
 Modifying fixed pattern into an adaptive way is a common idea to
 improve performance. There have been a great number of works
 help models focus on important features and adopt geometric variations in computer vision. All related works fall into two categories, attention-based methods
 \cite{senet,nonlocal,gcnet,ccnet} and offset-based methods \cite{dcn,dcnv2,deformabledetr,gfnet,racnn}. We mainly review offset-based ones.
 
 Offset-based methods predict offsets to explicitly direct important
 locations. This idea bears some similarity with region proposal
 network in object detection \cite{fasterrcnn,maskrcnn}. Unlike our task, region
 proposal network uses supervision of bounding box annotations.
 In image classification task, there are also some works explicitly learning positions
 of the important regions for better performance \cite{racnn} or faster inference \cite{gfnet}.
 The learning process are merely guided by cross-entropy loss and final accuracy. 
 Deformable convolution \cite{dcn,dcnv2} is the work most similar to ours. It predicts an offset for each pixel of the convolution kernel, while the predicted regions in our method are more regular ones. Irregular patches are not compatible in vision transformers. Deformable-DETR \cite{deformabledetr} applies deformable operations in the self-attention layers and cross-attention layers of DETR. However, its main purpose is to accelerate training, and Deformable-DETR still relies on feature maps extracted from CNNs. As far as we know, our work is the first to apply deformable operations in a pure vision transformer architecture. We focus on adjusting the position and scale of each patch, therefore extracting features better maintaining local structures. Our module can work as a plug-and-play module, and is compatible for various vision transformer architectures.

\section{Method}
\subsection{Preliminaries: Vision Transformer\label{sec:preliminary}}
Vision transformer is composed of three parts, a patch embedding module, multi-head self-attention blocks and feed-forward multi-layer perceptrons (MLP). The network starts with the patch embedding module which transforms the input image into a sequence of tokens, and then alternately stacks multi-head self-attention blocks and MLPs to obtain the final representation. We mainly elaborate on the patch embedding module in this section, and then have a quick review over the multi-head self-attention.

Patch embedding module divides images into patches with fixed size and positions, and embeds each of them with a linear layer. We denote the input image or feature map as $\bm{A}\in\mathbb{R}^{H\times W\times C}$. For simplicity, we assume $H=W$. Previous works split $\bm{A}$ into a sequence of $N$ patches with size $s\times s$ ($s=\lceil H/\sqrt{N}\rceil$). The sequence is denoted as $\{\bm{z}^{(i)}\}_{1\leq i\leq N}$. 

To better interpret the patch splitting process, we reformulate the patch embedding module. Each patch $\bm{z}^{(i)}$ can be seen as a representation for a rectangle region of the input image. We denote its center coordinate as $(x_{ct}^{(i)},y_{ct}^{(i)})$. Since patch size is fixed, the left-top corner and right-bottom corner are determined as $(x_{ct}^{(i)}-s/2,y_{ct}^{(i)}-s/2)$ and $(x_{ct}^{(i)}+s/2,y_{ct}^{(i)}+s/2)$. There are $s\times s$ pixels inside this region, their coordinates are represented by $\vec{\bm{p}}^{(i,j)}=(p_{x}^{(i,j)},p_{y}^{(i,j)})$. All coordinates $\vec{\bm{p}}^{(i,j)}$ are integers themselves. The features at these pixels are denoted as $\{\tilde{\bm{a}}^{(i,j)}\}_{1\leq j\leq N}$. These features are then flattened and processed by a linear layer to get representation for the new patch, as shown in Eq.~(\ref{equ:embedding}).
\begin{equation}
  \bm{z}^{(i)}=\bm{W}_{patch}\cdot concat\{\tilde{\bm{a}}^{(i,1)},...,\tilde{\bm{a}}^{(i,s\times s)}\}+\bm{b}_{patch}
  \label{equ:embedding}
\end{equation}

Multi-head self-attention module aggregates relative information over the whole input sequence, giving each token a global view. This module learns three groups of representative features for each head, query ($\bm{Q}_h\in\mathbb{R}^{N\times d}$), key ($\bm{K}_h\in\mathbb{R}^{N\times d}$) and value ($\bm{V}_h\in\mathbb{R}^{N\times d}$). $\bm{Q}_h$ and $\bm{K}_h$ are multiplied to obtain the attention map $\bm{Attn}_h$, which represents similarity between different patches. The attention map is used as the weights to sum up $\bm{V}_h$. Independent results are calculated for different heads to get more variant features. Results from all heads are then concatenated and transformed to become new representations $\bm{Z}'$.
\begin{equation}
  \bm{Attn}_h=Softmax(\bm{Q}_h\bm{K}_h^T/\sqrt{d})
\end{equation}
\begin{equation}
  \bm{Z}'=Concat\{\bm{Attn}_1\bm{V}_1, ..., \bm{Attn}_H\bm{V}_H\}\bm{W}_{proj}+\bm{b}_{proj}
\end{equation}

\subsection{DePatch Module \label{sec:module}}

\begin{figure}[h]
  \centering
  \subfigure[Vanilla patch embedding module in PVT]{
    \includegraphics[width=\linewidth]{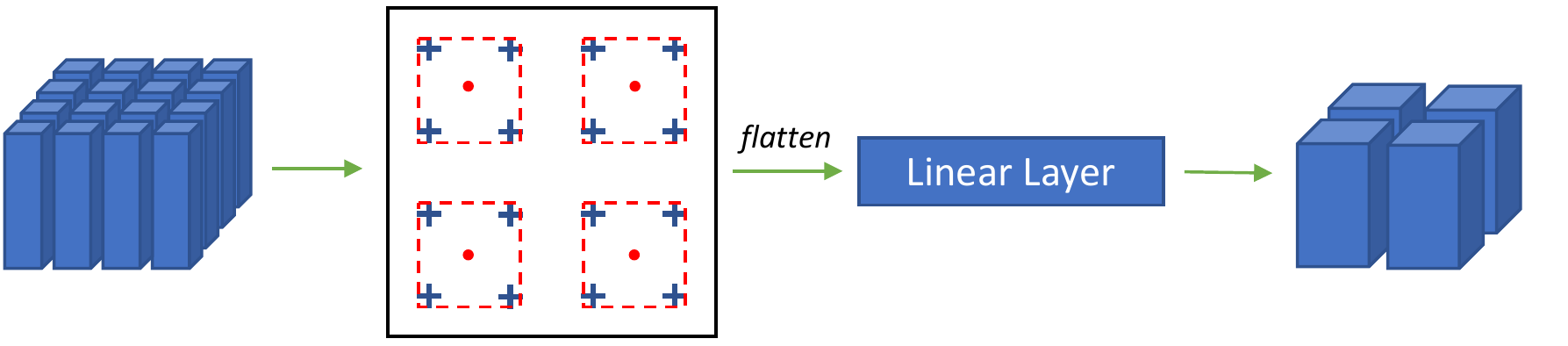}
    \label{fig:module1}
  }
  \centering
  \subfigure[DePatch module ($k=3$)]{
    \includegraphics[width=\linewidth]{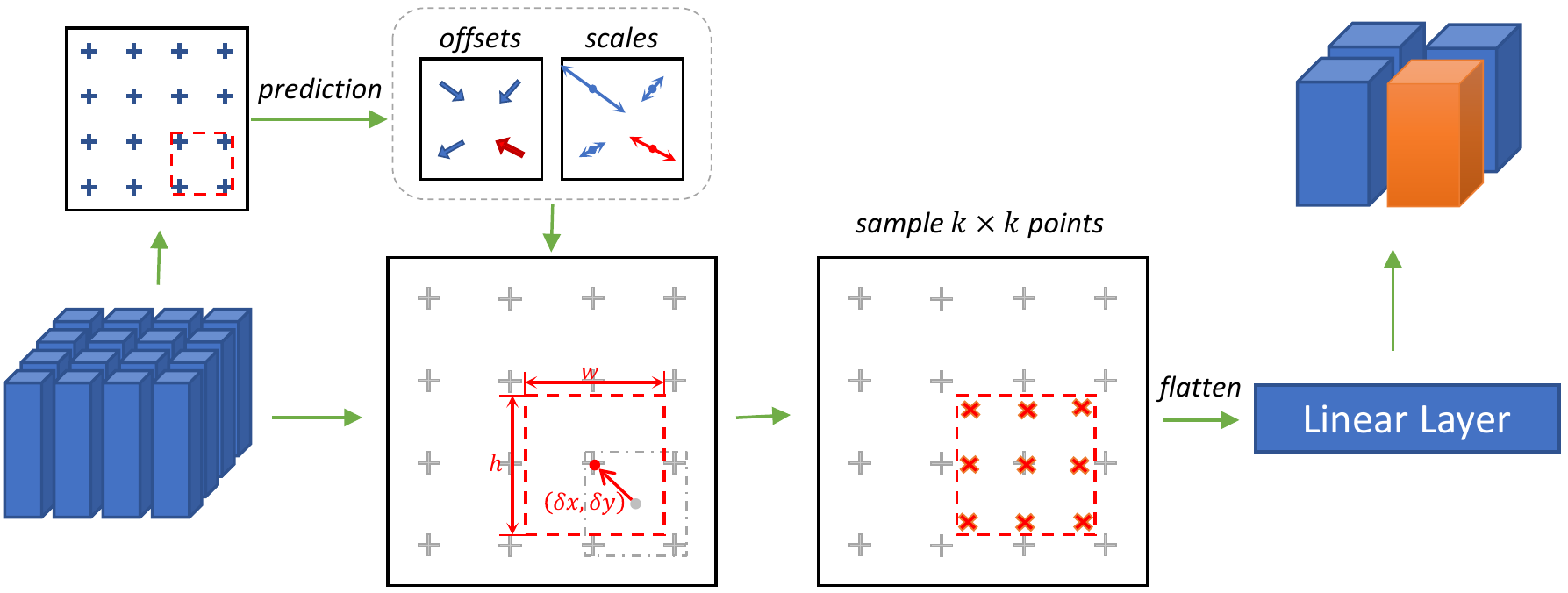}
    \label{fig:module2}
  }
  \caption{DePatch module instruction. Offsets and scales are predicted with local features, and new embeddings are obtained by bilinear interpolation.}
  \label{fig:DePatch}
  \Description{Our module design figure.}
\end{figure}

The patch embedding process described in \ref{sec:preliminary} is fixed and inflexible. Positions $(x_{ct}^{(i)},y_{ct}^{(i)})$ and size $s$ are fixed, therefore the rectangle region is unchangable for each patch. The feature for each patch is directly represented with its inside pixels. In order to better locate important structures and handle geometric deformation, we loosen these constraints to develop our deformable patch embedding module, DePatch.

Firstly, we turn the location and the scale of each patch into predicted parameters based on input contents. As for the location, we predict an offset $(\delta x, \delta y)$ allowing it to shift around the original center. As for the scale, we simply replace the fixed patch size $s$ with predictable $s_h$ and $s_w$. In this way, we can determine a new rectangle region, and denote its left-top corner as $(x_1, y_1)$ and right-bottom corner as $(x_2, y_2)$. For clarity, we omit the superscript $(i)$. We emphasize that $(\delta x, \delta y, s_w, s_h)$ can be different even for patches in a single image.
\begin{equation}
  \begin{split}
    x_1 = x_{ct}+\delta x - \frac{s_{w}}{2},\ y_1 = y_{ct}+\delta y - \frac{s_{h}}{2}\\
    x_2 = x_{ct}+\delta x + \frac{s_{w}}{2},\ y_2 = y_{ct}+\delta y + \frac{s_{h}}{2}
  \end{split}
  \label{equ:coordinate}
\end{equation}

As shown in Figure~\ref{fig:DePatch}, we add a new branch to predict these parameters. Based on the input feature map, we predict $(\delta x, \delta y, s_w, s_h)$ densely for all patches first, and then embed them with predicted regions. Offsets and scales are predicted with Eq.~(\ref{equ:pred_offset}) and (\ref{equ:pred_scale}). $\bm{f}_p(\cdot)$ can be any feature extractor, and here we use a single linear layer. After that, $\bm{W}_{offset}$ and $\bm{W}_{scale}$ are followed to predict offsets and scales. At the beginning of training, these weights are initialized to zero. $\bm{b}_{scale}$ is initialized to make sure each patch focuses on exactly the same rectangle region as the original model.
\begin{equation}
  \bm{\delta x},\ \bm{\delta y}=Tanh(\bm{W}_{offset}\cdot \bm{f}_p(\bm{A}))
  \label{equ:pred_offset}
\end{equation}
\begin{equation}
  \bm{s}_w, \bm{s}_h=ReLU(Tanh(\bm{W}_{scale}\cdot \bm{f}_p(\bm{A})+\bm{b}_{scale}))
  \label{equ:pred_scale}
\end{equation}
\begin{figure}[h]
  \centering
  \includegraphics[width=\linewidth]{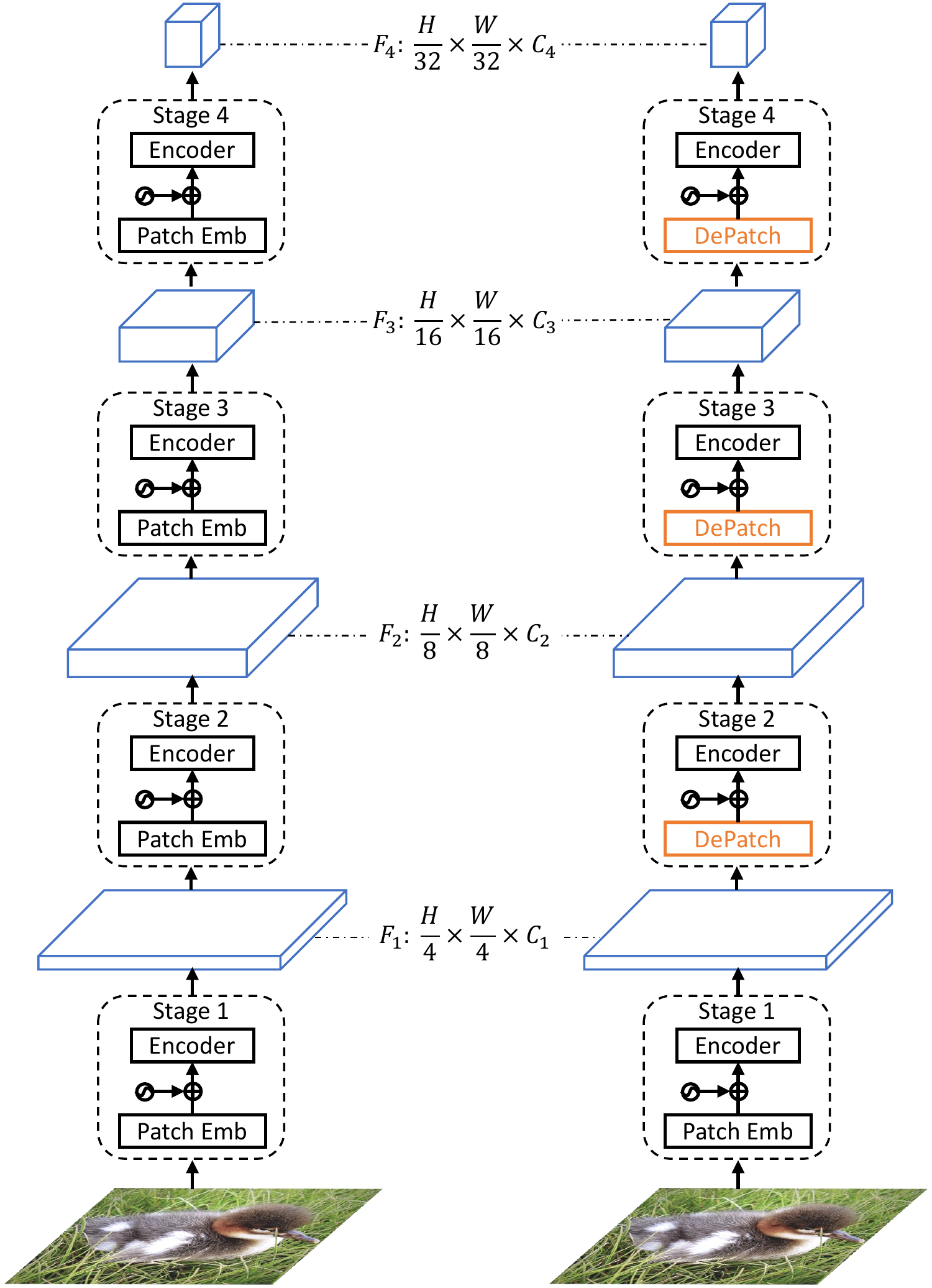}
  \caption{Left: Original PVT architecture. Right: DPT, Equipped with our DePatch module.}
  \label{fig:overall_arch}
  \Description{Our module design figure.}
\end{figure}

After the rectangle region is determined, we extract the feature for each patch. The main problem is that regions are with different sizes, and the predicted coordinates are usually fractional. We solve this problem in a sampling-and-interpolation manner. Given the rectangle coordinates $(x_1, y_1)$ and $(x_2, y_2)$, we sample $k\times k$ points uniformly inside the region, $k$ is a super-parameter for our method, which will be ablated in \ref{sec:ablation}. Each sampling location is denoted as $\vec{\bm{p}}^{(j)}=(p_{x}^{(j)},p_{y}^{(j)})$ for any $1\leq j\leq k\times k$. The features of all sampled points $\{\tilde{\bm{a}}^{(j)}\}_{1\leq j\leq k\times k}$ are then flattened and processed in a linear layer to generate patch embedding, as in Eq.~(\ref{equ:projection}).
\begin{equation}
  \bm{z}^{(i)}=\bm{W}\cdot concat\{\tilde{\bm{a}}^{(1)}, ...,\tilde{\bm{a}}^{(k\times k)}\}+\bm{b}
  \label{equ:projection}
\end{equation}

The index of sampled points is mostly fractional. Assume that we intend to extract feature at point $(p_x, p_y)$. Its corresponding feature is obtained via bilinear interpolation as
\begin{equation}
  \bm{A}(p_{x}, p_{y})=\sum_{q_{x},q_{y}}{G(p_{x},p_{y};q_{x},q_{y})\cdot \bm{A}(q_{x}, q_{y})}
  \label{equ:bilinear}
\end{equation}
\begin{equation}
  G(p_{x}, p_{y};q_{x},q_{y})=max(0,1-|p_{x}-q_{x}|)\cdot max(0,1-|p_{y}-q_{y}|)
\end{equation}

In Eq.~(\ref{equ:bilinear}), $G(\cdot)$ is the bilinear interpolation kernel all over the integral spatial locations. It only becomes non-zero at the four locations close to $(p_x, p_y)$. Therefore, it can be computed quickly with few multiply-adds.

\subsection{Overall Architecture}

DePatch is a self-adaptive module to change positions and scales of patches. 
As DePatch can work as a plug-and-play module, we can easily incorporate DePatch into various vision transformers. 
Because of the superiority and generality, we choose PVT as our base model. PVT has four stages with feature maps of decreasing scales. It utilizes spatial-reduction attention to reduce cost on high resolution feature maps. For detailed information please refer to \cite{pvt}. Our model is denoted as DPT. It is built by replacing the patch embedding modules at the beginning of stage 2, 3 and 4 with DePatch, while keeping other configurations unchanged. The overall architecture is shown in Figure~\ref{fig:overall_arch}.

\section{Experiments}
In this section, we conduct image classification experiments on ImageNet \cite{imagenet} and object detection experiments on COCO \cite{coco}. After that, some ablation studies are provided then for further analysis.

\subsection{Image Classification}

\textbf{Experiment Settings} We use ImageNet \cite{imagenet} for image classification experiments. The ImageNet dataset consists of 1.28M images for training and 50K for validation. These images belong to 1000 categories. We report top-1 error on validation set for comparison. The images are resized into $256\times 256$ and randomly cropped into $224\times 224$ for training. Advanced data augmentation methods including Mixup \cite{mixup}, CutMix \cite{cutmix}, label smoothing \cite{labelsmooth} and Rand-Augment \cite{randaug} are utilized. Our models are trained with the batch size of 1024 for 300 epochs and optimized by AdamW \cite{adamw} with initial learning rate of $1\times10^{-3}$ and cosine schedule \cite{cosine}. Weight decay is set to 0.05 for non-bias parameters. All these settings keep the same with original PVT \cite{pvt} for fair comparison.

\textbf{Results} As shown in Table~\ref{tab:cls_result}, our smallest DPT-Tiny achieves 77.4\% top-1 accuracy, which is 2.3\% higher than the corresponding baseline PVT model. The best result is achieved by our medium one. It achieves 81.9\% top-1 accuracy, and even outperforms models with much larger costs like PVT-Large and catch up with DeiT-Base. As for CNN-based models, DPT-Small outperforms the popular ResNet50 by 4.9\%. Our models achieve better results than both CNN-based and transformer-based models, and outperform them by a large margin.

\begin{table}
  \caption{Results on ImageNet Classification.}
  \label{tab:cls_result}
  \begin{tabular}{lccc}
    \toprule
    Method & \#Param (M) & FLOPs (G) & Top-1 Acc(\%)\\
    \midrule
    ResNet18 \cite{resnet}  & 11.7 & 1.8 & 69.8\\
    DeiT-Tiny \cite{deit}   & 5.7  & 1.3 & 72.2\\
    PVT-Tiny \cite{pvt}     & 13.2 & 1.9 & 75.1\\
    \textbf{DPT-Tiny (ours)}& 15.2 & 2.1 & \textbf{77.4}\\
    \midrule
    ResNet50 \cite{resnet}   & 25.6 & 4.1 & 76.1\\
    DeiT-Small \cite{deit}   & 22.1 & 4.6 & 79.9\\
    T2T-ViT-14 \cite{t2tvit} & 21.4 & 5.2 & 80.6\\
    PVT-Small \cite{pvt}     & 24.5 & 3.8 & 79.8\\
    \textbf{DPT-Small (ours)}& 26.4 & 4.0 & \textbf{81.0}\\
    \midrule
    ResNet101 \cite{resnet}   & 44.7 & 7.9 & 77.4\\
    X101-32x4d \cite{resnext} & 44.2 & 8.0 & 78.8\\
    X101-64x4d \cite{resnext} & 83.5 & 15.6& 79.6\\
    ViT-Base \cite{vit}       & 86.6 & 17.6& 77.9\\
    DeiT-Base \cite{deit}     & 86.6 & 17.6& 81.8\\
    T2T-ViT-19 \cite{t2tvit}  & 39.0 & 8.0 & 81.2\\
    PVT-Medium \cite{pvt}     & 44.2 & 6.7 & 81.2\\
    PVT-Large \cite{pvt}      & 61.4 & 9.8 & 81.7\\
    \textbf{DPT-Medium (ours)}& 46.1 & 6.9 & \textbf{81.9}\\
  \bottomrule
\end{tabular}
\end{table}

\subsection{Object Detection}
\begin{table*}
  \caption{Object detection performance on MS COCO (RetinaNet)}
  \label{tab:det_retina_result}
  \begin{tabular}{l|c|cccccc|cccccc}
    \toprule
     & & \multicolumn{6}{c}{RetinaNet 1$\times$} & \multicolumn{6}{c}{RetinaNet 3$\times$ + MS}\\
    Backbone & \#Param(M) & $mAP$ & $AP_{50}$ & $AP_{75}$ & $AP_S$ & $AP_M$ & $AP_L$ & $AP$ & $AP_{50}$ & $AP_{75}$ & $AP_S$ & $AP_M$ & $AP_L$\\
    \midrule
    ResNet18 \cite{resnet}& 21.3 & 31.8 & 49.6 & 33.6 & 16.3 & 34.3 & 43.2 & 35.4 & 53.9 & 37.6 & 19.5 & 38.2 & 46.8\\
    PVT-Tiny \cite{pvt}   & 23.0 & 36.7 & 56.9 & 38.9 & 22.6 & 38.8 & 50.0 & 39.4 & 59.8 & 42.0 & 25.5 & 42.0 & 52.1\\
    \textbf{DPT-Tiny (ours)}       & 24.9 & \textbf{39.5} & \textbf{60.4} & \textbf{41.8} & \textbf{23.7} & \textbf{43.2} & \textbf{52.2} & \textbf{41.2} & \textbf{62.0} & \textbf{44.0} & \textbf{25.7} & \textbf{44.6} & \textbf{53.9}\\
    \midrule
    ResNet50 \cite{resnet}& 37.7 & 36.3 & 55.3 & 38.6 & 19.3 & 40.0 & 48.8 & 39.0 & 58.4 & 41.8 & 22.4 & 42.8 & 51.6\\
    PVT-Small \cite{pvt}  & 34.2 & 40.4 & 61.3 & 43.0 & 25.0 & 42.9 & 55.7 & 42.2 & 62.7 & 45.0 & 26.2 & 45.2 & 57.2\\
    \textbf{DPT-Small (ours)}      & 36.1 & \textbf{42.5} & \textbf{63.6} & \textbf{45.3} & \textbf{26.2} & \textbf{45.7} & \textbf{56.9} & \textbf{43.3} & \textbf{64.0} & \textbf{46.5} & \textbf{27.8} & \textbf{46.3} & \textbf{58.5}\\
    \midrule
    ResNet101 \cite{resnet}        & 56.7  &  38.5 & 57.8 & 41.2 & 21.4 & 42.6 & 51.1  &  40.9 & 60.1 & 44.0 & 23.7 & 45.0 & 53.8\\
    ResNeXt101-32x4d \cite{resnext}& 56.4  &  39.9 & 59.6 & 42.7 & 22.3 & 44.2 & 52.5  &  41.4 & 61.0 & 44.3 & 23.9 & 45.5 & 53.7\\
    ResNeXt101-64x4d \cite{resnext} & 95.5 &  41.0 & 60.9 & 44.0 & 23.9 & 45.2 & 54.0  &  41.8 & 61.5 & 44.4 & 25.2 & 45.4 & 54.6\\
    PVT-Medium \cite{pvt}          & 53.9  &  41.9 & 63.1 & 44.3 & 25.0 & 44.9 & 57.6  &  43.2 & 63.8 & 46.1 & 27.3 & 46.3 & \textbf{58.9}\\
    PVT-Large \cite{pvt}            & 71.1 &  42.6 & 63.7 & 45.4 & 25.8 & 46.0 & 58.4  &  43.4 & 63.6 & 46.1 & 26.1 & 46.0 & 59.5\\
    \textbf{DPT-Medium (ours)}     & 55.9  &  \textbf{43.3} & \textbf{64.6} & \textbf{45.9} & \textbf{27.2} & \textbf{46.7} & \textbf{58.6}  &  \textbf{43.7} & \textbf{64.6} & \textbf{46.4} & \textbf{27.2} & \textbf{47.0} & 58.4\\
  \bottomrule
\end{tabular}
\end{table*}

\begin{table*}
  \caption{Object detection performance on MS COCO (Mask R-CNN)}
  \label{tab:det_mask_result}
  \begin{tabular}{l|c|cccccc|cccccc}
    \toprule
    & & \multicolumn{6}{c}{Mask R-CNN 1$\times$} & \multicolumn{6}{c}{Mask R-CNN 3$\times$ + MS}\\
    Backbone & \#Param(M) & $mAP^b$ & $AP^b_{50}$ & $AP^b_{75}$ & $mAP^m$ & $AP^m_{50}$ & $AP^m_{75}$ & $mAP^b$ & $AP^b_{50}$ & $AP^b_{75}$ & $mAP^m$ & $AP^m_{50}$ & $AP^m_{75}$\\
    \midrule
    ResNet18 \cite{resnet}  & 31.2  & 34.0 & 54.0 & 36.7 & 31.2 & 51.0 & 32.7 & 36.9 & 57.1 & 40.0 & 33.6 & 53.9 & 35.7\\
    PVT-Tiny \cite{pvt}     & 32.9 & 36.7 & 59.2 & 39.3 & 35.1 & 56.7 & 37.3 & 39.8 & 62.2 & 43.0 & 37.4 & 59.3 & 39.9\\
    \textbf{DPT-Tiny (ours)}& 34.8 & \textbf{40.2} & \textbf{62.8} & \textbf{43.8} & \textbf{37.7} & \textbf{59.8} & \textbf{40.4} & \textbf{42.2} & \textbf{64.4} & \textbf{46.1} & \textbf{39.4} & \textbf{61.5} & \textbf{42.3}\\
    \midrule
    ResNet50 \cite{resnet}  & 44.2  & 38.0 & 58.6 & 41.4 & 34.4 & 55.1 & 36.7 & 41.0 & 61.7 & 44.9 & 37.1 & 58.4 & 40.1\\
    PVT-Small \cite{pvt}    & 44.1 & 40.4 & 62.9 & 43.8 & 37.8 & 60.1 & 40.3 & 43.0 & 65.3 & 46.9 & 39.9 & 62.5 & 42.8\\
    \textbf{DPT-Small (ours)}      & 46.1 & \textbf{43.1} & \textbf{65.7} & \textbf{47.2} & \textbf{39.9} & \textbf{62.9} & \textbf{43.0} & \textbf{44.4} & \textbf{66.5} & \textbf{48.9} & \textbf{41.0} & \textbf{63.6} & \textbf{44.2}\\
    \midrule
    ResNet101 \cite{resnet} & 63.2  & 40.4 & 61.1 & 44.2 & 36.4 & 57.7 & 38.8 & 42.8 & 63.2 & 47.1 & 38.5 & 60.1 & 41.3\\
    ResNeXt101-32x4d \cite{resnext} & 62.8 & 41.9 & 62.5 & 45.9 & 37.5 & 59.4 & 40.2 & 44.0 & 64.4 & 48.0 & 39.2 & 61.4 & 41.9\\
    ResNeXt101-64x4d \cite{resnext} & 101.9 & 42.8 & 63.8 & 47.3 & 38.4 & 60.6 & 41.3 & 44.4 & 64.9 & 48.8 & 39.7 & 61.9 & 42.6\\
    PVT-Medium \cite{pvt}           & 63.9 & 42.0 & 64.4 & 45.6 & 39.0 & 61.6 & 42.1 & 44.2 & \textbf{66.0} & 48.2 & 40.5 & \textbf{63.1} & 43.5\\
    PVT-Large \cite{pvt}            & 81.0  & 42.9 & 65.0 & 46.6 & 39.5 & 61.9 & 42.5 & 44.5 & 66.0 & 48.3 & 40.7 & 63.4 & 43.7\\
    \textbf{DPT-Medium (ours)}      & 65.8 & \textbf{43.8} & \textbf{66.2} & \textbf{48.3} & \textbf{40.3} & \textbf{63.1} & \textbf{43.4} & \textbf{44.3} & 65.6 & \textbf{48.8} & \textbf{40.7} & \textbf{63.1} & \textbf{44.1}\\
    \midrule
  \bottomrule
\end{tabular}
\end{table*}

\begin{table}
  \caption{Object detection performance on MS COCO (DETR with 50 epochs)}
  \label{tab:det_detr_result}
  \begin{tabular}{lccccccc}
    \toprule
    Backbone & $mAP$ & $AP_{50}$ & $AP_{75}$ & $AP_S$ & $AP_M$ & $AP_L$\\
    \midrule
    ResNet50 \cite{resnet}   & 32.3 & 53.9 & 32.3 & 10.7 & 33.8 & 53.0\\
    PVT-Small \cite{pvt}     & 34.7 & 55.7 & 35.4 & 12.0 & 36.4 & 56.7\\
    \textbf{DPT-Small (ours)}& \textbf{37.7} & \textbf{59.2} & \textbf{38.8} & \textbf{15.0} & \textbf{40.3} & \textbf{58.5}\\
  \bottomrule
\end{tabular}
\end{table}

\textbf{Experiment Settings} Our experiments for object detection are conducted on COCO \cite{coco}, a large-scale detection benchmark. We set train2017 split with 118K images as our training set, and val2017 split with 5K images for validation. Mean Average Precision (mAP) is used as our evaluation metric. Following \cite{pvt}, we evaluate our DPT backbones on three prevailing frameworks, RetinaNet \cite{retina}, Mask R-CNN \cite{maskrcnn} and DETR \cite{detr}. We load ImageNet pretrained weights to initialize the backbone. Our models are trained with the batch size of 16 and optimized by AdamW \cite{adamw} with initial learning rate $1\times10^{-4}$. As to RetinaNet and Mask R-CNN, we report results with both 1x and multi-scale 3x train schedules (12 and 36 epochs). For 1x schedule, images are resized so that the shorter edge has 800 pixels and the longer edge does not exceed 1333 pixels. For multi-scale training, the shorter edge is resized within the range of [640, 800]. As for DETR, the model is trained for 50 epochs with random flip and random scale.

\textbf{Results} We compare DPT to PVT \cite{pvt} and standard ResNe(X)t \cite{resnet,resnext}. The comparison is shown in Table~\ref{tab:det_retina_result}. As for RetinaNet, DPT-Small significantly outperforms PVT-Small by 2.1\% and Resnet50 by 6.2\% mAP at a comparable computational cost, which indicates that DPT provides more discriminative features for target objects in images. With our DePatch module, each patch is aware of its neighboring content, and extracts crucial information needed for different locations. Moreover, with $3\times$ training schedule and multi-scale training, RetinaNet+DPT-Medium achieves 43.7\% mAP. It outperforms PVT-Medium and ResNet101 by a large margin, and even achieves better performance than PVT-Large model, but with more than 20\% cost reduced. 

Results for Mask R-CNN are similar. Our DPT-Small model achieves 43.1\% box mAP and 39.9\% mask mAP under $1\times$ schedule, outperforming PVT-Small by 2.7\% and 2.1\%. With $3\times$ training schedule and multi-scale training, Mask R-CNN+DPT-Small achieves the best result with 44.4\% box mAP and 41.0\% mask mAP.

DETR is a latest framework for object detection. It requires a long trained schedule (e.g. 500 epochs). We only validate our model with a shorter schedule (50 epochs). According to Table~\ref{tab:det_detr_result}, our DPT-Small achieves 37.7\% box mAP, outperforming PVT-Small by 3.0\% and ResNet50 by 5.4\%. Therefore we conclude that DPT is also compatible with transformer-based detectors.

\subsection{Ablation Studies\label{sec:ablation}}
\begin{figure}[h]
  \centering
  \subfigure[Scale distribution at stage 2]{
    \includegraphics[width=\linewidth]{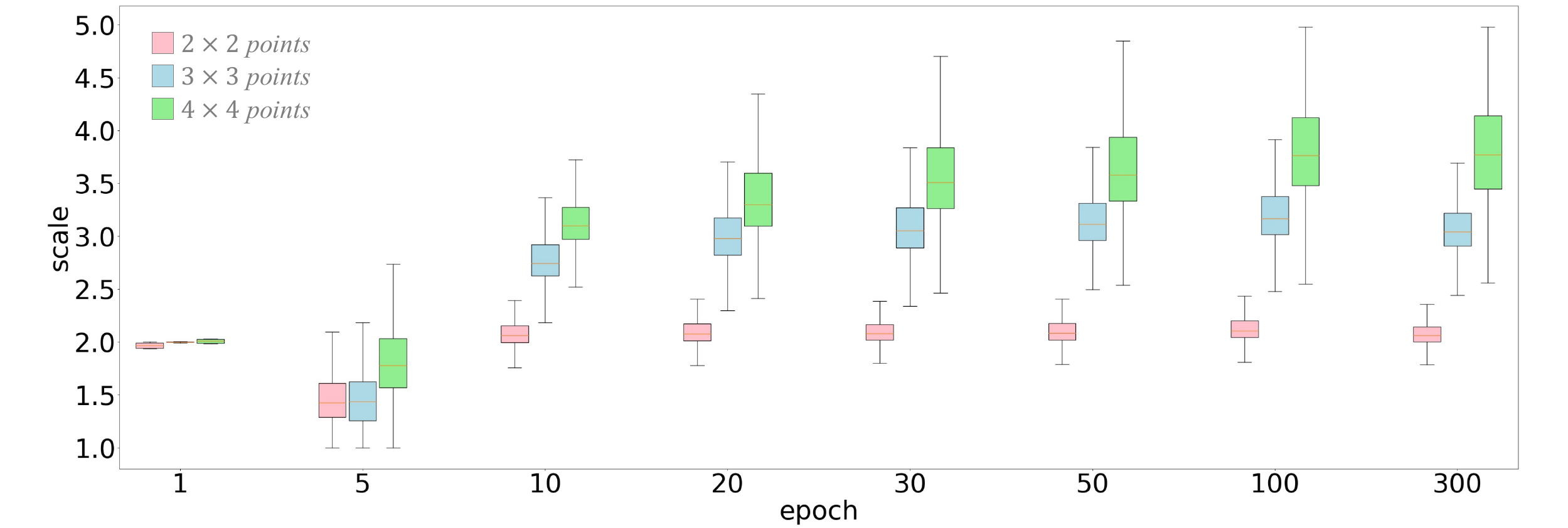}
  }
  \centering
  \subfigure[Scale distribution at stage 3]{
    \includegraphics[width=\linewidth]{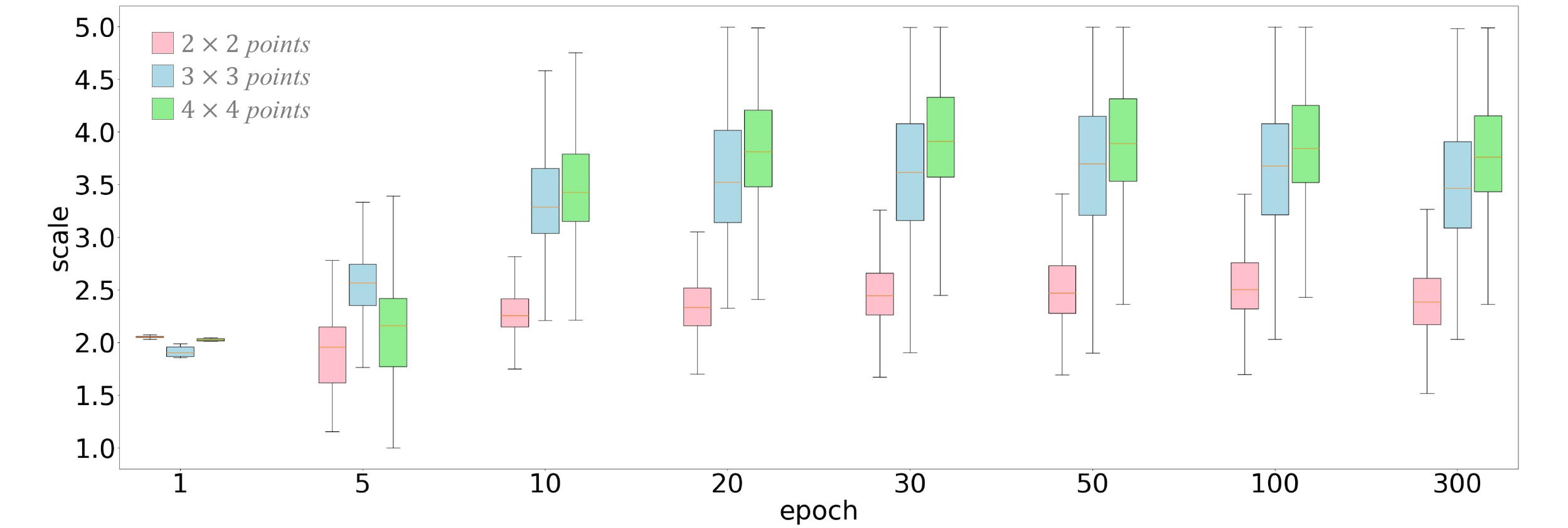}
  }
  \centering
  \subfigure[Scale distribution at stage 4]{
    \includegraphics[width=\linewidth]{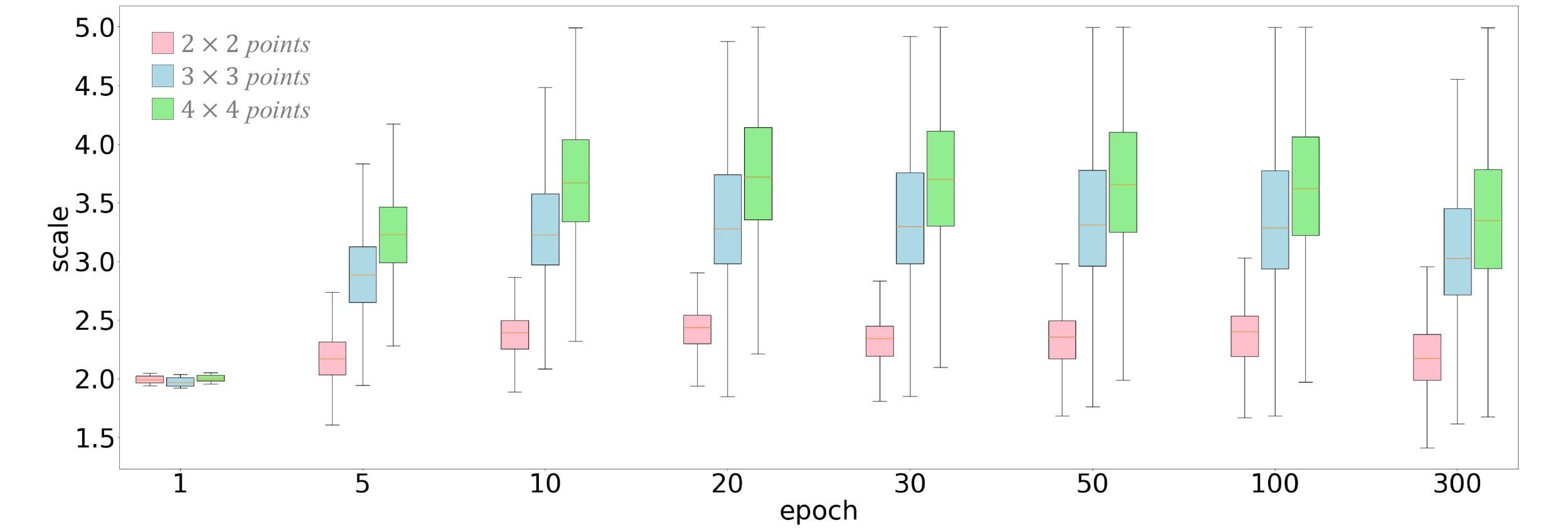}
  }
  \caption{Region scale learned by DPT-Tiny with different number of sampling points $k\times k$. We illustrate statistics in stage 2, 3 and 4. Scale is measured by the edge size of the region.}
  \label{fig:analyze_scale}
  \Description{Region scale learned by DPT-Tiny.}
\end{figure}
\textbf{Effect of module position}
There are four patch embedding modules in PVT. The first directly operates on the input image, and the rest are inserted at the beginning of the following stages. We perform detailed experiments to illustrate where we should add DePatch.

Since raw images contain little semantic information, it is difficult for the first module to predict offsets and scales beyond its own region. Therefore, we only attempt to replace the rest three patch embedding modules. The results are shown in Table~\ref{tab:add_position}. The improvements obtained by stage 2, 3 and 4 are 0.3\%, 1.0\%, and 1.5\%. The more patch embedding modules we replace, larger improvement it brings. According to the results, replacing all patch embedding modules in stage 2, 3 and 4 achieves the best performance. It outperforms baseline PVT model by 1.5\%, with only 0.86M parameters increased. In the following experiments, it will be our default configuration to replace all three patch embedding modules.
\begin{table}
  \caption{Effect of module position ($k=2$)}
  \label{tab:add_position}
  \begin{tabular}{ccccc}
    \toprule
    Stage 2 & Stage 3 & Stage 4 & \#Params(M) & Top-1 Acc(\%)\\
    \midrule
               &            &            & 13.23 & 75.1\\
    \checkmark &            &            & 13.26 & 75.4\\
    \checkmark & \checkmark &            & 13.43 & 76.1\\
    \checkmark & \checkmark & \checkmark & 14.09 & 76.6\\
  \bottomrule
\end{tabular}
\end{table}
\begin{table}
  \caption{Effect of number of sampling points}
  \label{tab:sampling_points}
  \begin{tabular}{cccc}
    \toprule
    Sampling points & \#Params(M) & Flops(G) & Top-1 Acc(\%)\\
    \midrule
    Baseline  & 13.23 & 1.94 & 75.1\\
    \midrule
    $2\times 2$ & 14.09 & 2.03 & 76.6\\
    $3\times 3$ & 15.15 & 2.14 & 77.4\\
    $4\times 4$ & 16.64 & 2.30 & 77.6\\
  \bottomrule
\end{tabular}
\end{table}
\begin{table}
  \caption{Decouple of predicting offsets and scales}
  \label{tab:offset_scale}
  \begin{tabular}{ccc}
    \toprule
    Offsets & Scales & Top-1 Acc(\%)\\
    \midrule
               &            & 75.1\\
    \checkmark &            & 76.6\\
    \checkmark & \checkmark & 77.4\\
  \bottomrule
\end{tabular}
\end{table}
\begin{table}
  \caption{Top-1 accuracy (\%) with short training schedule}
  \label{tab:fast_train}
  \begin{tabular}{lcc}
    \toprule
    Method &  150 epochs & 300 epochs\\
    \midrule
    PVT-Tiny \cite{pvt} & 73.1 & 75.1\\
    DPT-Tiny (Ours)     & 76.2 & 77.4\\
  \bottomrule
\end{tabular}
\end{table}

\textbf{Effect of number of sampling points}
We experiment to show how many points we should sample in one predicted region. Sampling more points slightly increases FLOPs, but also has a stronger learning ability to capture features from a larger region. The results are shown in Table~\ref{tab:sampling_points}. Increasing sampling points from $2\times 2$ to $3\times 3$ provides another 0.8\% improvement, while further increasing it to $4\times 4$ only improves by $0.2\%$. Since sampling $4\times 4$ points only gets marginal improvement. We take the $k=3$ as the default configuration in following experiments.

\begin{figure}[h]
  \subfigure[]{
    \includegraphics[width=0.46\linewidth]{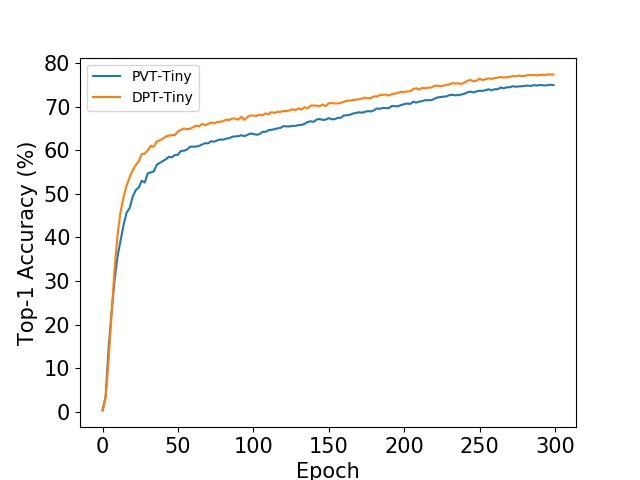}
  }
  \subfigure[]{
    \includegraphics[width=0.46\linewidth]{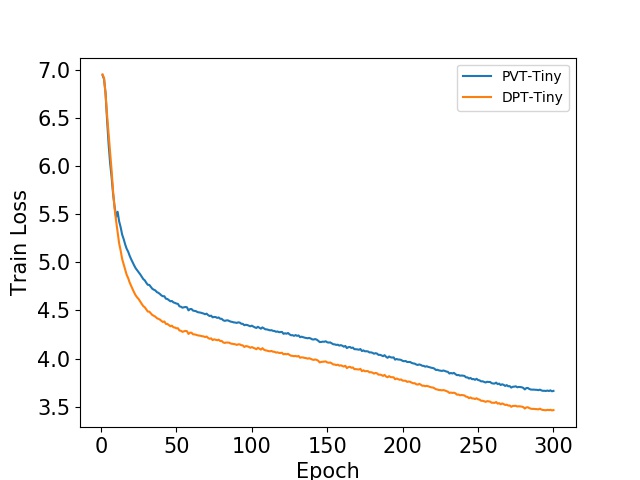}
  }
  \caption{Training curve for both DPT-Tiny and PVT-Tiny.}
  \label{fig:training_curve}
  \Description{Region scale learned by DPT-Tiny.}
\end{figure}

We claim that sampling more points will benefits DPT with stronger ability to extract features from larger area. We show how the distributions of predicted scales change during training with different number of sampling points in Figure~\ref{fig:analyze_scale}. Although we initialize the region scales strictly the same as that in PVT ($patch\_size=2$), DePatch can learn to expand on its own. This phenomenon accords to the common sense in CNN, that enlarging the receptive field benefits the model. The distributions of scales for $k=2$ converge early with low variance. Sampling $2\times 2$ points is unable to represent any larger regions, hence it limits the capacity of the our module for understanding images with heavy geometric deformation. The statistics do not differ much for $k=3$ and $4$ except in stage 2. We assume that sampling more points will achieve even better performance, while it is not worth additional cost. Designing a more sophisticated spatial pyramid cound be another way to improve our method. We leave it as our future work.

\textbf{Decouple offsets and scales}
DePatch learns both the offset and scale for each patch. Offsets are predicted to shift the patches towards more important regions, and scales are for better maintaining local structures. They both facilitates the performance of our model. We decouple these two factors in Table~\ref{tab:offset_scale} in order to see how each single one influences our model. When scales are not predicted, the shape of all rectangle regions is fixed the same as patches in original PVT. Only predicting offsets can improve 1.5\% above baseline, and another 0.8\% is obtained by predicting scales. We claim that both offsets and scales are important for our self-adaptive patch embedding module.

\textbf{Analysis for fast convergence}
DePatch module is able to adjust the patches to a proper shape for each image. Adequate patches maintain important local structures, and features are learned more efficiently. Therefore the whole network can learn at a faster speed. We draw the training curve for both our DPT-Tiny and PVT-Tiny in Figure~\ref{fig:training_curve}. The training loss and test accuracy do converge faster in first few epochs.

Based on this phenomenon, we expect that our module can alleviate the requirement of long training schedule. We prove it by simply reducing training epochs by half. As shown in Table~\ref{tab:fast_train}, DPT-Tiny trained with only 150 epochs outperforms a fully-trained PVT-Tiny by 1.1\%, and the performance degradation caused by a shorter schedule is only 1.2\%, which is much smaller than original PVT-Tiny. This indicates that our DePatch module can significantly accelerate training for vision transformers, which would benefit further research.

\textbf{Effect of parameter initialization}
As stated in \ref{sec:module}, we initialize $W_{offset}$ and $W_{scale}$ to zero. We shown in Table~\ref{tab:result_init} that initialization methods have little impact on final performance. We take zero initialization in the all experiments.
\begin{table}
  \caption{Effect of initialization for $W_{offset}$ and $W_{scale}$}
  \label{tab:result_init}
  \begin{tabular}{cc}
    \toprule
    Initialization & Top-1 Acc(\%)\\
    \midrule
    Truncated normal & 77.36\\
    Zero init        & 77.39\\
  \bottomrule
\end{tabular}
\end{table}

\subsection{Visualization}
\begin{figure*}[h]
  \subfigure[]{
    \includegraphics[width=0.23\linewidth]{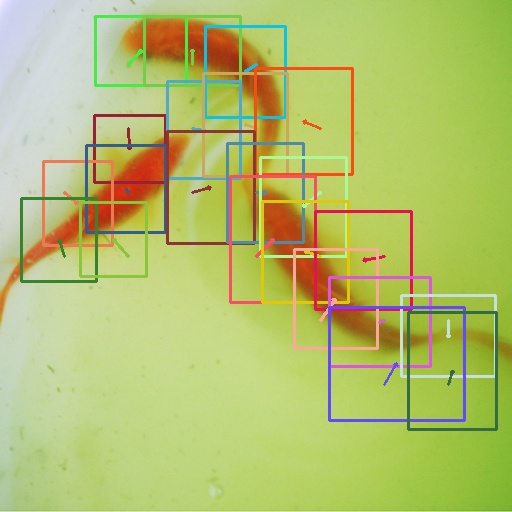}
  }
  \subfigure[]{
    \includegraphics[width=0.23\linewidth]{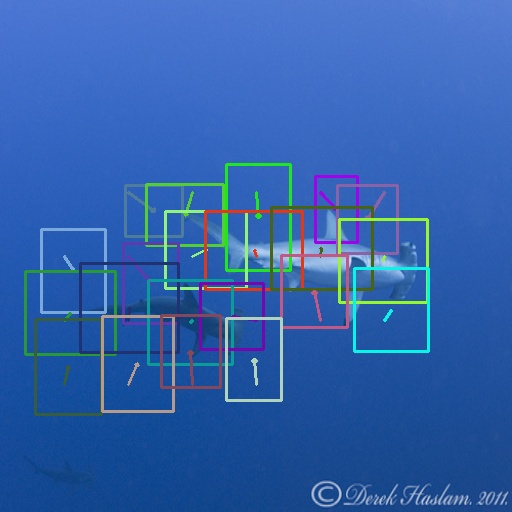}
    \label{fig:visual_whale}
  }
  \subfigure[]{
    \includegraphics[width=0.23\linewidth]{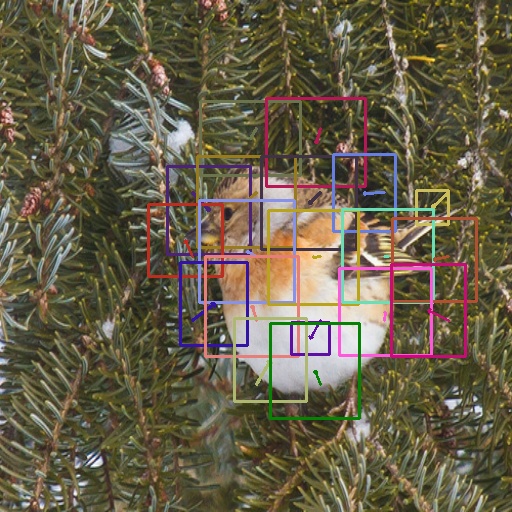}
  }
  \subfigure[]{
    \includegraphics[width=0.23\linewidth]{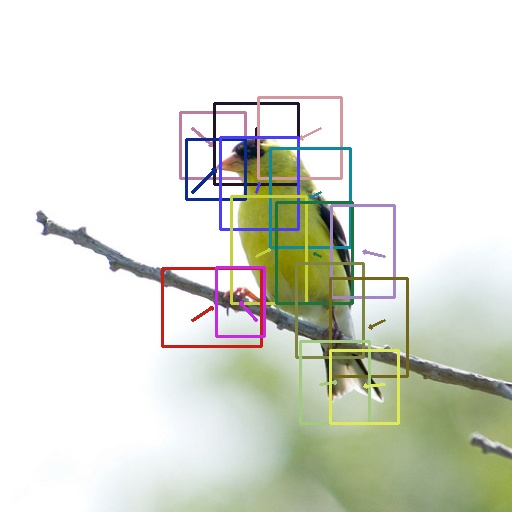}
    \label{fig:visual_bird}
  }
  \subfigure[]{
    \includegraphics[width=0.23\linewidth]{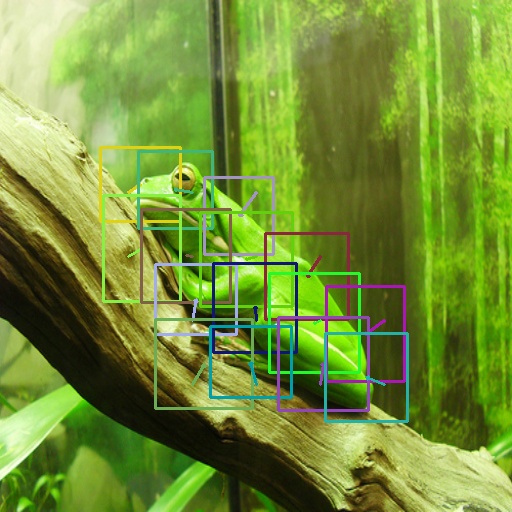}
  }
  \subfigure[]{
    \includegraphics[width=0.23\linewidth]{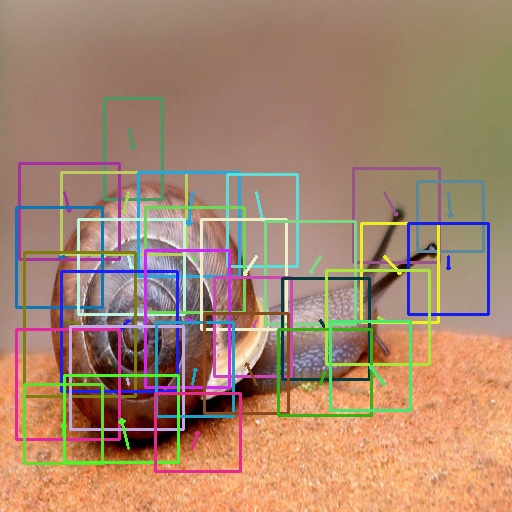}
  }
  \subfigure[]{
    \includegraphics[width=0.23\linewidth]{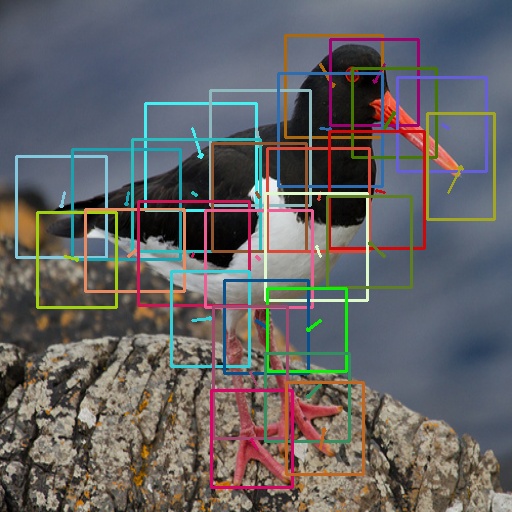}
  }
  \subfigure[]{
    \includegraphics[width=0.23\linewidth]{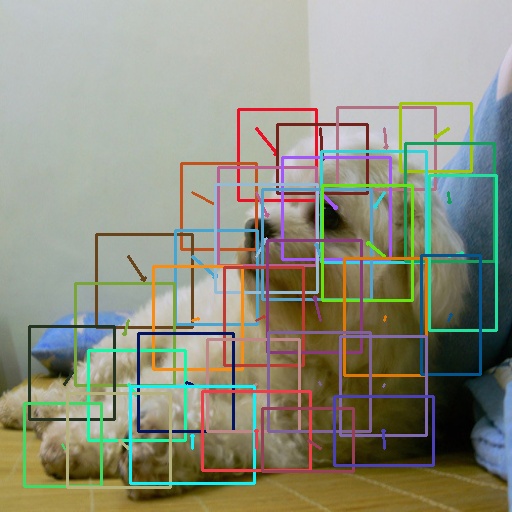}
    \label{fig:visual_dog}
  }
  \caption{Visualization of learned patches and their offsets with our DPT-Small at stage 4. Our method can adaptively adjust the position and scale for each patch based on the input content.}
  \label{fig:offset_imagenet}
  \Description{Visualization figure.}
\end{figure*}
We illustrate offsets learned by DePatch in Figure~\ref{fig:offset_imagenet}. The visualization shows that patches predicted by DePatch are well-located to capture important features. DePatch has more obvious impacts at the edge of foreground objects. It encourages the patches outside to shift a bit towards the object, thus covering more critical areas than normal patches. When there are more than one object appearing in the image, patches would adjust their positions to the closest one (the two whales in Figure~\ref{fig:visual_whale}). This attribute would be more crucial for object detection, since different patches can be more representative for different objects. Therefore, the detector can better locate and classify all the objects with more related features. The predicted scales are also influenced by richness of local context, such as edges or corners. It becomes small when it needs to focus on subtle details (the beak of the bird in Figure~\ref{fig:visual_bird}), and large if more context is needed (homogenous area of the dog's stomach Figure~\ref{fig:visual_dog}). The high variance of offsets and scales indicates strong self-adaptability of our method.

\section{Conclusion}
In this paper, we introduce DePatch, a deformable module to split patches. It encourages the network to extract patch information from object-related regions and make our model insensitive to geometric deformation. This module can work as a plug-and-play module and improve various vision transformers. We also build a transformer with DePatch module, named Deformable Patch-based Transformer, DPT. Extensive experiments on image classification and object detection indicate that DPT can extract better features and outperform CNN-based models and other vision transformers. DePatch can be utilized in other vision transformers as well as other downstream tasks to improve their performance. Our model is the first work to modify vision transformer in a data-dependant way. We hope our idea could serve as a good starting point for future studies.

\begin{acks}
This work was supported by the Research and Development Projects in the Key Areas of Guangdong Province (No.2020B010165001) and National Natural Science Foundation of China under Grants No.61772527, No.61976210, No.62002357, No.61876086, No.61806200, No.62002356, and No.61633002.
\end{acks}

\bibliographystyle{ACM-Reference-Format}
\bibliography{main-base}

\end{document}